\documentclass[conference]{IEEEtran}
\IEEEoverridecommandlockouts
\usepackage{hyperref}
\usepackage{amsmath,amssymb,amsfonts}

\usepackage{algorithmic}
\usepackage{graphicx}
\usepackage{textcomp}
\usepackage{xcolor}
\usepackage[finalizecache=false,frozencache=true]{minted}
\usepackage{tikz}
\usepackage{dcolumn}
\usetikzlibrary{shapes.misc, positioning}
\usepackage[subrefformat=parens]{subcaption}
\def\BibTeX{{\rm B\kern-.05em{\sc i\kern-.025em b}\kern-.08em
    T\kern-.1667em\lower.7ex\hbox{E}\kern-.125emX}}
\newcommand\hig{Higgins et al.~\cite{DARLA}}
\newcommand\zhang{Zhang et al.~\cite{Zhang1804}}
\newcommand\zhangns{Zhang et al.} 
\newcommand\cobbe{Cobbe et al.~\cite{Coinrun}}
\newcommand\cobbens{Cobbe et al.} 
\newcommand\juli{Juliani et al.~\cite{ObsTower}}

\newcommand\fare{Farebrother et al.~\cite{GeneralizeDQN}}
\newcommand\macha{Machado et al.~\cite{Machado18}}
\newcommand\sco{stochasticity}
\newcommand\sym{\texttt}
\newcommand\ppos{$\textsc{PPO Small}$}
\newcommand\ppol{$\textsc{PPO Large}$}
\newcommand\ppob{$\textsc{PPO BatchNorm}$}
\newcommand\bvae{$\beta$-VAE}
\newcommand\bppo{\bvae~PPO}
\newcommand\pcg{procedural content generation} 

\newcommand\MDP[1]{
$\mathcal{M}_#1=(\mathcal{S}_#1,\mathcal{A}_#1,\mathcal{R}_#1,\mathcal{P}_#1)$%
}
\newcommand\footurl[1]{\footnote{\url{#1}~(Accessed:2019-05-29)}}
\setminted{fontsize=\small,baselinestretch=1}
\begin{document}
\title{Rogue-Gym: A New Challenge for Generalization in Reinforcement Learning
}

\author{\IEEEauthorblockN{Yuji Kanagawa}
\IEEEauthorblockA{\textit{Graduate School of Arts and Sciences} \\
\textit{The University of Tokyo}\\
Tokyo, Japan \\
kanagawa-yuji968@g.ecc.u-tokyo.ac.jp}
\and
\IEEEauthorblockN{Tomoyuki Kaneko\thanks{A part of this work was supported by JSPS KAKENHI
Grant Number 18K19832 and by JST, PRESTO.}
}
\IEEEauthorblockA{\textit{Interfaculty Initiative in Information Studies} \\
\textit{The University of Tokyo}\\
Tokyo, Japan \\
kaneko@graco.c.u-tokyo.ac.jp}
}
\IEEEpubid{\begin{minipage}{\textwidth}\ \\[12pt]
978-1-7281-1884-0/19/\$31.00 \copyright 2019 IEEE
\end{minipage}}
\maketitle
\begin{abstract}
 In this paper, we propose Rogue-Gym, a simple and classic style roguelike
 game built for evaluating generalization in reinforcement
 learning~(RL).
 Combined with the recent progress of deep neural networks, RL has
 successfully trained human-level agents without human knowledge in many
 games such as those for Atari 2600.
 However, it has been pointed out that agents trained with RL methods
 often overfit the training environment, and they work poorly in slightly
 different environments.
 To investigate this problem, some research environments with \pcg{}
 have been proposed.
 Following these studies, we propose the use of roguelikes as a
 benchmark for evaluating the generalization ability of RL agents.
 In our Rogue-Gym, agents need to explore dungeons that are
 structured differently each time they start a new game.
 Thanks to the very diverse structures of the dungeons, we believe that
 the generalization benchmark of Rogue-Gym is sufficiently fair.
 In our experiments, we evaluate a standard reinforcement learning
 method, PPO, with and without enhancements for generalization.
 The results show that some enhancements believed to be effective fail to
 mitigate the overfitting in Rogue-Gym, although others slightly
 improve the generalization ability.
\end{abstract}


\section{Introduction} \label{intro}
Reinforcement learning~(RL) is a key method for training
AI agents without human knowledge.
Recent advances in deep reinforcement learning have created human-level
agents in many games, such as those for Atari 2600~\cite{DQN} and
the game DOOM~\cite{MnihBMGLHSK16}, using only pixels as inputs.
This method could be applied to many domains, from robotics to the game
industry.

However, it is still difficult to generalize learned policies between
tasks even for current state of the art RL algorithms.
Recent studies (e.g., by \zhang{} and by \cobbe{}) have shown that agents
trained by RL methods often overfit the \textit{training environment}
and perform poorly in a \textit{test environment}, when the test
environment is not exactly the same as the training environment.
This is an important problem because test environments are often
differ somewhat from training environments in many applications of
reinforcement learning.
For example, in real world applications including self-driving
cars~\cite{DriveInaDay}, agents are often trained via simulators or
designated areas but need to perform safely in real world situations
that are similar to but different from their training environments.
For agents to act appropriately in unknown situations, they need to
properly generalize their policies that they learned from the training
environment.
Generalization is also important in transfer learning, where the goal is
to transfer a policy learned in a training environment to another
similar environment, called the \textit{target environment}.
We can use this method to reduce the training time in many applications
of RL.
For example, we can imagine a situation in which we train an enemy in an
action game through experience across fewer stages and then transfer 
the enemy to a higher number of other scenes via generalization.

In this paper, we propose the Rogue-Gym environment, a simple roguelike
game built to evaluate the generalization ability of RL agents.
As in the original implementation of Rogue, it has very diverse and
randomly generated dungeon structures on each floor.
Thus, there is no pattern of actions that is always effective, which
makes the generalization benchmark in Rogue-Gym sufficiently fair.
Instead, in Rogue-Gym agents have to generalize abstract subgoals like
getting coins or going downstairs through their action sequences
that consist of concrete actions (e.g., moving left).
Rogue-Gym is designed so that the environment an agent encounters,
which includes dungeon maps, items, and enemies, is configurable through
a random seed.
Thus, we can easily evaluate the generalization score in Rogue-Gym by
using random seeds different from those used in training.
Since many other properties including the size of dungeons and the
presence of enemies, are completely configurable, researchers can easily
adjust the difficulty of learning so that the properties are complex
enough and difficult for simple agents to solve but can still be
addressed by the state-of-the-art RL methods.

In our experiments, we evaluate a popular DRL algorithm with or without
generalization methods in Rogue-Gym.
We show that some of the methods work poorly in generalization, although they
successfully improve the training scores through learning.
In contrast, some of these methods, like L2 regularization, achieve better
generalization scores than those of the baseline methods, but
the results are not sufficiently effective. 
Therefore, Rogue-Gym is a novel and challenging domain for further studies. 

\section{Background}\label{section:background}
We follow a standard notation and denote a Markov decision process
$\mathcal M$ by $\mathcal{M} = (\mathcal{S,A,R,P})$, where $\mathcal S$
is the state space, $\mathcal A$ is the action space, $\mathcal R$ is
the immediate reward function that maps a state to a reward, and
$\mathcal P$ is the state transition probability.
We denote the policy of an agent by $\pi(a|s)$,
that is, the probability of taking an action
$a\in\mathcal{A}$ given a state $s\in\mathcal{S}$.
In an instance of MDP, the goal of reinforcement
learning~\cite{SuttonBarto2018}(RL) is to get the optimal policy that
maximizes the expected total reward by repeatedly taking an action,
observing a state, and getting a reward.
In this paper, we consider the episodic setting, in which the total
reward is defined as $R = \sum_{t=0}^T \mathcal{R}(s_t)$, where $t$
denotes the time step, and the initial state $s_0$ is sampled from the
initial state distribution $\mathcal{P}^0$.

One of the famous classes of RL algorithms is policy gradients. Suppose
that a policy $\pi$ is parameterized by a parameter vector $\theta$. 
Then, we can denote the policy by $\pi_\theta$ and the gradient of the
expected sum of the reward by $\nabla_\theta \mathbb{E}[R]$.
The goal of policy gradient methods is to maximize $\mathbb{E}[R]$ by
iteratively updating $\theta$ on the basis of estimating of
$\nabla_\theta \mathbb{E}[R]$ with agents' experience.

Deep reinforcement learning refers to RL methods that use deep
neural networks as function approximators.
DRL enables us to train RL agents given only screen pixels as states,
through the use of deep convolutional neural networks (CNNs)~\cite{DQN}.
PPO~\cite{PPO} is one of the state-of-the-art deep policy gradient
methods, and we use it as a baseline method in this paper.

\section{Related Work} \label{section:related-work}
\fare{} proposed the use of ALE~\cite{ALE}, an environment based on
an Atari2600 emulator, to evaluate generalization in RL.
They conducted experiments by using different game modes of Atari 2600
games introduced by \macha{} and showed that regularization
techniques like L2 regularization mitigates the overfitting of
DQN~\cite{DQN}.
However, the number of environments is limited in ALE, which allows us
to tune algorithms for specific environments.

To increase the number of training/evaluation environments, \pcg{} is
considered to be a promising method.
\zhang{} conducted experiments by using simple 2D gridworld mazes generated
procedurally and showed that some of the typical methods used for mitigating
overfitting in RL often fail.

\cobbe{} proposed the CoinRun environment, which procedurally generates
short 2D action games that have different backgrounds and stage
structures.
They showed that large neural network architectures and standard
regularization methods such as batch normalization~\cite{IoffeS15}
help policy generalization in CoinRun. 
In addition, it is notable that both \zhangns{} and \cobbens{} reported
that increasing the number of training levels helps generalization.

\juli{} proposed Obstacle Tower, where the player explores procedurally
generated 3D dungeons from a third person perspective.
Inspired by \textit{Montezuma's Revenge}, one of the most difficult
games that can be played in ALE, they designed Obstacle Tower to include
factors like sparse rewards, which makes the task hard for RL algorithms.
In experiments, they showed that the state-of-the-art algorithms
including PPO struggle to generalize learned policies in Obstacle Tower.

Our work is most similar to \cobbe{} and \juli{} in proposing a new
environment with \pcg{} for evaluating generalization, though we place
more importance on customizability and reasonable difficulty.

In addition to regularization, state representation
learning~\cite{SRL} is also a promising approach for generalization.
The key idea is the use of abstract state representations to bridge the
gap between a training environment and a test environment. We can
obtain such a representation via unsupervised learning
methods such as variational autoencoders~(VAE)~\cite{KingmaW13}.

\hig{} adopted this idea for RL and proposed DARLA, which learns
disentangled state representations by using \bvae{}~\cite{betaVAE} from
randomly collected observations and then learns a policy by using these
representations. They manually set up training and test environments by
changing the colors and/or locations of objects in 3D navigation tasks in
DeepMind Lab~\cite{DMLab}. They showed that DARLA improves generalization
scores in these tasks.
We evaluated \bvae{} in our experiments.

\section{Rogue-Gym Environment}\label{section:rogue-gym}
In this section, we introduce the Rogue-Gym environment, which is a
simple roguelike game built for evaluating the generalization performance
of RL agents.

To fairly evaluate the generalization ability of RL algorithms, we claim
that structural diversity across training and test environments is
important, in addition to a sufficient number of test environments.
In the context of evaluating an RL agent in a single task, \macha{}
claimed that the \sco{} of an environment is important by showing that a
simple algorithm that memorizes only an effective sequence of actions
performs well in a deterministic environment.
This kind of hack is also possible in a generalization setting if the
training and test environments do not have diverse structures and share
an undesirable structural similarity.
For example, in the \textit{normal} task of CoinRun~\cite{Coinrun},
stages share a common structure in that the player is initially on the left
side of the stage, and the coin of each stage is placed on the right
side.
This means that we can perform well by always moving or jumping
to the right in almost all stages.
In fact, we observed that a random agent that selects only
\texttt{right} and \texttt{right-jump} completed about the $62$\% of
stages.

On the basis of this claim, we propose the use of roguelikes as a testbed for
generalization.
In this paper, we use the term \textit{roguelike} as a subgenre of
role-playing video games, where a player explores procedurally generated
dungeons\footnote{Note that this definition is popular and consistent
with the description on
Wikipedia~\url{https://en.wikipedia.org/wiki/Roguelike}}.
Our Rogue-Gym is a variant of roguelike and inherits the following
properties desirable for evaluating generalization:
\begin{enumerate}
 \item it is naturally integrated with procedural generation and provides
       us with a lot of test environments,
 \item it has several behaviors agents need to generalize, such as finding
       doors or fighting enemies, and
 \item it has very diverse dungeon structures, which prevents memorizing
       hacks.
\end{enumerate}

In addition to these crucial properties, the following conditions are also important
for enabling research on various learning methods with various computing resources available:
\begin{enumerate}
 \item easy to customize and
 \item able to change the difficulty with sufficient granularity.
\end{enumerate}
We believe that customizability is especially important since the
learning time required by deep RL algorithms heavily depends on the
screen size.

To satisfy all of the properties, we create and present a simple roguelike
named Rogue-Gym, with a clean implementation made from scratch by the first
author, following the behavior of the original Rogue as accurately as
possible.
This is because other roguelike games popular for human players, such as NetHack and
Cataclysm: DDA, are often too complex for RL agents.
Also, the implementation of the original Rogue was written in old style
C and is hard to modify.

\begin{figure}[t]
 \centering
 \includegraphics[width=5cm]{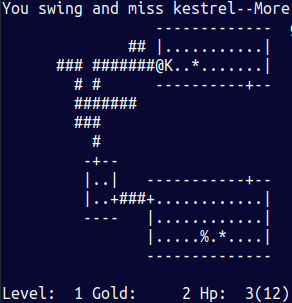}
 \caption{Screenshot of Rogue-Gym} \label{ss1}
\end{figure}

Fig.~\ref{ss1} shows a screenshot of Rogue-Gym.
Like many roguelike games, it has a command line interface based on ASCII
characters.  Table~\ref{syms} summarizes the meanings of characters. 
\begin{table}[t]
  \centering
 \caption{Meaning of characters} \label{syms}
  \begin{tabular}{l|l} \hline
   Character & Meaning \\ \hline
   \sym{@} & Player \\
   \sym{.} & Floor \\
   \sym{\#} & Passage \\
   \sym{|}, \sym{-} & Wall \\
   \sym{*} & Gold \\
   \sym{\%} & Downstairs \\
   \sym{+} & Door \\ 
   \sym{A}-\sym{Z} & Enemy (disabled in experiments in Sect.~\ref{section:experiments})
\\ \hline
  \end{tabular}
\end{table}

\begin{figure}[t]
  \begin{minipage}[cbt]{4cm}
   \centering
   \includegraphics[width=3.0cm]{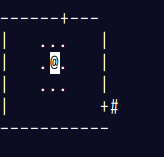}
   \subcaption{Dark room}
  \end{minipage}
  \begin{minipage}[cbt]{4cm}
   \centering
   \includegraphics[width=3.0cm]{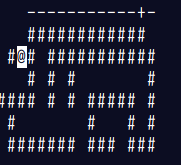}
   \subcaption{Maze room}
  \end{minipage}
 \caption{Example of room variants} \label{room}
\end{figure}

In Rogue-Gym, the mission of the player is to get the Amulet of Yendor
hidden on the deepest floor by finding the way to get downstairs on
each floor.
One floor consists of several rooms and passages but still has many
combinatorial patterns, which makes it desirable for evaluating
generalization.
As shown in Fig.~\ref{room}, in addition to normal rooms, Rogue-Gym has

\begin{figure}[t]
  \begin{minipage}[cbt]{4cm}
   \centering
   \includegraphics[width=2.5cm]{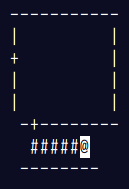}
   \subcaption{Before \texttt{search}}
  \end{minipage}
  \begin{minipage}[cbt]{4cm}
   \centering
   \includegraphics[width=2.5cm]{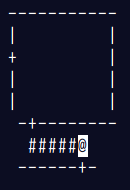}
   \subcaption{After \texttt{search}}
  \end{minipage}
  \caption{Example of hidden door: 
 a door exists but is not observable (left). With (several uses of)
 the search command, the door was discovered (right).} \label{hidden}
\end{figure}

There are dark rooms, in which only nine grids around the player are visible,
and maze rooms consisting of passages that are arranged in a complicated manner.
Rogue-Gym also has a variety of transition dynamics.
\begin{figure}[t]
  \begin{minipage}[cbt]{3cm}
   \centering
   \includegraphics[width=2cm]{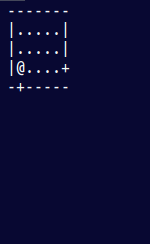}
   \subcaption{Initial state}
  \end{minipage}
  \begin{minipage}[cbt]{6cm}
    \centering
    \includegraphics[width=2cm]{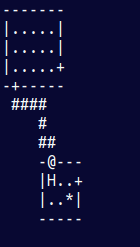}
    \includegraphics[width=2cm]{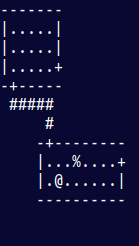}
   \subcaption{Possible future states}
  \end{minipage}
 \caption{The same observation may lead to different future states;
   while there is a door to
   the south of the player, the rest of the map is hidden before
   the door is opened (left).  The agent may be faced with different states 
   depending on the random seed after passing through the door (right).}
 \label{states}
\end{figure}
Since only the areas that an agent has visited are visible by default,
there can be situations where current states are the same but future
states are different depending on the random seed, as shown in
Fig.~\ref{states}.
In addition, Rogue-Gym is partially observable because of hidden
passages and doors. As shown in Fig.~\ref{hidden},
in Rogue-Gym, passages and doors are sometimes hidden and block the
player's way.  In this situation, if the player uses the \texttt{search}
command, passages and doors can appear at a certain probability. \par

Rogue-Gym is available at the GitHub
repository\footurl{https://github.com/kngwyu/rogue-gym}.
It is written in Rust in order to speed up performance but designed so
that it can be called from many programming languages.
The Python API is the main interface for training AI agents, of which 
binary packages are available at
PyPI\footurl{https://pypi.org/project/rogue-gym/} and installable via
\texttt{pip install rogue\_gym}.
As shown by the example code in Fig.~\ref{api}, the API allows users to
configure Rogue-Gym flexibly via Python's \texttt{dict} or JSON. We can
change the size of the screen, kinds of enemies and items, and so on.
\begin{figure}[t]\small
 \begin{minted}{python}
from rogue_gym.envs import RogueEnv
CONFIG = {
    'width': 32, 'height': 16,
    'dungeon': {
        'style': 'rogue',
        'room_num_x': 2, 'room_num_y': 2
     }
}
env = RogueEnv(max_steps=100, config_dict=CONFIG)
rewards = 0
state = env.reset()
for i in range(10): 
    # move right
    state, reward, done, _ = env.step('l')
    rewards += reward
 \end{minted}
 \caption{Python API example of Rogue-Gym} \label{api}
\end{figure}
The \texttt{RogueEnv} class inherits the \texttt{Env} class of OpenAI
Gym~\cite{gym}, a standard library for defining RL environments. 
Thus, we believe our Rogue-Gym is compatible with much existing RL code
and easy to use.
An observation for an agent is encoded as an image having the same
number of binary channels as the number of ASCII characters, following a
standard procedure (e.g., Silver et al.~\cite{AlphaGo}).
This encoding is suitable for the input of a convolutional neural
network (CNN).

\section{Generalization Problem in Rogue-Gym} \label{section:problem-setting}
In this section, we describe a concrete problem in Rogue-Gym that can
be used to evaluate the generalization ability of RL algorithms.
Here, our goal of generalization is to obtain abstract strategies, such
as exploring new passages and rooms, from a limited number of samples.

To make the notations clear, we consider generalization in RL as a form
of domain adaptation~\cite{GlorotBB11} problems where the target domain
is unknown.
Following a study by \hig{}, we define domain adaptation in RL as
a problem in which we train agents in the source MDP\MDP{S} with
$\mathcal{P}^0_S$, and then evaluate them in the target MDP\MDP{T} with
$\mathcal{P}^0_T$. Though the set of states $\mathcal S_{S}$ and
$\mathcal S_{T}$ can be different, we assume that the action spaces
$\mathcal A_{S}$ and $\mathcal A_{T}$ are the same and that reward
functions $\mathcal R_{S},\, \mathcal R_{T}$ and transition
$\mathcal P_{S},\, \mathcal P_{T}$ share some underlying structure.
This problem setting looks different from but is essentially the same as
those by \zhang{} or by \cobbe{}. \par

In the experiments in this paper, we define the source environment
$(\mathcal{M}_S, \mathcal{P}^0_S)$ as s set of random number seeds, and
also define the target environment $(\mathcal{M}_T, \mathcal{P}^0_T)$ as
a different set of random number seeds that does not overlap with the
source ones.
Since the number of target environments is sufficiently large, we treat
the score for the target environment as the generalization score.

Configurations other than random seeds for dungeons are the same in the
source and target domains. 
In all experiments, we configured the screen size to be $32\times 16$
without enemies and items.
Each episode ended at 500 time steps.
This configuration makes the problem not too difficult and sufficiently
easy to learn.
The action space was discrete with 11 dimensions, as listed in 
Table.~\ref{rogue-actions}.
\begin{table}[t]
  \centering
  \caption{All actions used in experiments} \label{rogue-actions}
  \begin{tabular}{l|c}\hline
   Command & Meaning \\ \hline
   \texttt{.} & No operation \\
   \texttt{h} & Move left \\
   \texttt{j} & Move up \\
   \texttt{k} & Move down \\
   \texttt{l} & Move right \\
   \texttt{n} & Move right down \\
   \texttt{b} & Move left up \\
   \texttt{u} & Move right up \\
   \texttt{y} & Move left down \\
   \texttt{>} & Go downstairs \\
   \texttt{s} & Search around player \\ \hline
  \end{tabular}
\end{table}

Rewards consist of gold that an agent gathers throughout an episode.
In addition, we give $50$ golds as pseudo rewards each time an agent
reaches the next floor to adjust the difficulty to be suitable for
standard RL algorithms.
Hence, reward functions $\mathcal{R}_{S}$ and $\mathcal{R}_{T}$
are different due to the difference in the state space, but they still
share an underlying structure in that an agent can get a reward when:
\begin{itemize}
 \item it arrives at a grid with \sym{*}, and
 \item it arrives at a grid with \sym{\%} and selects \sym{downstairs}
       as action.
\end{itemize}
Transition probabilities $\mathcal{P}_{S, T}$ are also different, but they
share an underlying game rule that indicates how the player can move in
dungeons, which includes \sco{} like that shown in Fig.~\ref{states}.

As metric of generalization ability, we simply use the average value of
rewards in all target environments, which can be denoted by
$\mathbb{E}_{\pi}\left[\frac{1}{|\mathcal{M}_T|} \sum_{m\in \mathcal{M}_T} \sum_{t=0}
\mathcal{R}_m(s_t) \right]$.
In the following sections, we call this metric the
\textit{generalization score}.

\section{Evaluating Generalization Methods}\label{section:experiments}
In this section, we use Rogue-Gym to evaluate several methods used for
deep reinforcement learning and discuss their performance by using the
generalization score defined in the previous section.
Our code for the experiments is available at the GitHub
repository\footurl{https://github.com/kngwyu/rogue-gym-agents-cog19}.

\subsection{Reinforcement Learning Methods}
We used PPO~\cite{PPO} as our baseline because it was the best among
popular RL algorithms including DQN~\cite{DQN} and
A2C~\cite{MnihBMGLHSK16,ACKTR} in our preliminary experiments.
We compare the following six enhancements to encourage generalization on
top of PPO.
\begin{enumerate}
 \item \ppos{}: PPO with small CNN
 \item \ppol{}: PPO with large CNN
 \item \ppob{}: PPO with large CNN and batch normalization
 \item PPO L2: PPO with large CNN and L2 regularization
 \item VAE PPO: PPO with VAE
 \item \bppo{}: PPO with \bvae{}
\end{enumerate}
PPO Small uses a network with three CNN layers, which is similar to the
one used in DQN~\cite{DQN}.
PPO Large uses a network that is almost the same as the one used in
IMPALA~\cite{IMPALA}, which has 15 CNN layers and 6 residual
connections.
We adopted this large architecture because it was effective in a study by
\cobbe{}.
The hyper-parameters of PPO were mostly taken from the study~\cite{PPO}
and listed in Table.~\ref{ppo-params}.

\begin{table}[t]
  \centering
  \caption{PPO parameters} \label{ppo-params}
  \begin{tabular}{l|D{.}{.}{-1}} \hline
    Number of workers $N$ & 32 \\
    Rollout length & 125 \\
    Value coef. & 0.5 \\
    Entropy coef. & 0.01 \\
    $\gamma$ & 0.99 \\
    GAE $\lambda$ & 0.95 \\
    Num. epochs & 10 \\
    Clipping parameter $\epsilon$ & 0.1 \\
    Minibatch size & 200 \\
    Learning rate of Adam & \multicolumn{1}{c}{2.5e-4} \\
    $\epsilon$ of Adam & \multicolumn{1}{c}{1.0e-4} \\ \hline
 \end{tabular}
\end{table}

As standard regularization methods, we adopted batch normalization
(\textsc{PPO BatchNorm}) and L2 regularization (PPO L2).  Batch
normalization performed best among the three regularization methods used
by \cobbe{}, and L2 regularization is reported to improve regularization
in RL both by \cobbe{} and \fare{}.
We used $10^{-4}$ as a weight decay parameter $\lambda$ of PPO L2.

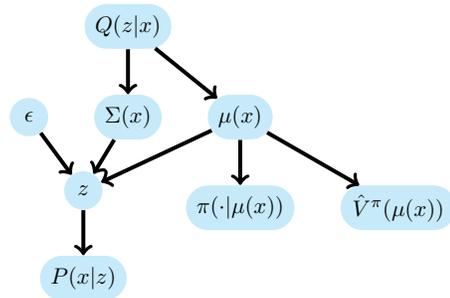
\begin{figure}[t]
 \centering
 \begin{tikzpicture}[node distance=5mm,scale=0.85,
                     var/.style={scale=0.85,
                         circle,
                         fill=cyan!20,
                         minimum height=6mm},
                     fn/.style={scale=0.85,
                         rounded rectangle,
                         fill=cyan!20,
                         minimum height=7mm}]
    \node[fn](q) {$Q(z|x)$};
    \node[fn, below=.6cm of q](s) {$\Sigma(x)$};
    \node[fn, right=.6cm of s](m) {$\mu(x)$};
    \node[fn, below=.6cm of m](pi) {$\pi(\cdot|\mu(x))$};
    \node[fn, right=.6cm of pi](v) {$\hat{V}^\pi(\mu(x))$};
    \node[var, left=.6cm of s](e) {$\epsilon$};
    \node[var, below right=of e, yshift=-3mm](z) {$z$};
    \node[fn, below=.6cm of z](p) {$P(x|z)$};
    \foreach \u / \v in {z/p,m/z,e/z,s/z,q/m,q/s,m/pi,m/v}
        \draw[->, line width=1.5pt] (\u) -- (\v);
 \end{tikzpicture}
 \caption{Computation graph of \bppo{}, where $x$ is the input, $z$ is a
 latent variable, $\pi$ is a policy, $\hat{V}$ is baseline used for the
 policy gradient, and $\mu$ and $\sigma$ are the mean and variance of a
 Gaussian distribution.} \label{proposed-graph}
\end{figure}

VAE PPO and \bppo{} were adopted to evaluate the effectiveness of
disentangled state representation learning, which was shown to be useful
in a study by \hig{}.
We adopted \bppo{} in a simpler manner than their DARLA. 
For simplicity in our implementation, reconstruction loss was calculated
without denoising autoencoders, while two versions, with and without
denoising autoencoders, were used for DARLA.
In addition, our \bppo{} trains \bvae{} simultaneously with a
policy by using parameter sharing, while DARLA trains \bvae{} with
random actions before learning a policy.
This is because it is difficult in Rogue-Gym to obtain sufficiently
diverse observations with only random actions.
We show a computation graph of \bppo{} in Fig.~\ref{proposed-graph}.

VAE PPO is a special case of \bppo{}~\cite{betaVAE}, where $\beta=0$.
It is used for comparison with $\beta > 0$ cases, which learns
disentangled representations.
We used $\beta = 4.0$ for \bppo{}.
Also, we added the score of the random agent to the bottom line.

As a training score, we used episodic rewards averaged over total
$1\,000$ trials ($100$ episodes for each of $10$ seeds $[0, 9])$) for the
random agent and used rewards averaged over $1000$ consecutive episodes
in the training time for other agents.
As generalization scores, we used rewards averaged over total $10\,000$
episodes ($10$ episodes for each of $1\,000$ seeds) for all agents.

\subsection{Difference in Training and Generalization Scores}
\begin{figure}[t]
  \begin{minipage}[cbt]{8cm}
   \centering
   \includegraphics[width=8cm]{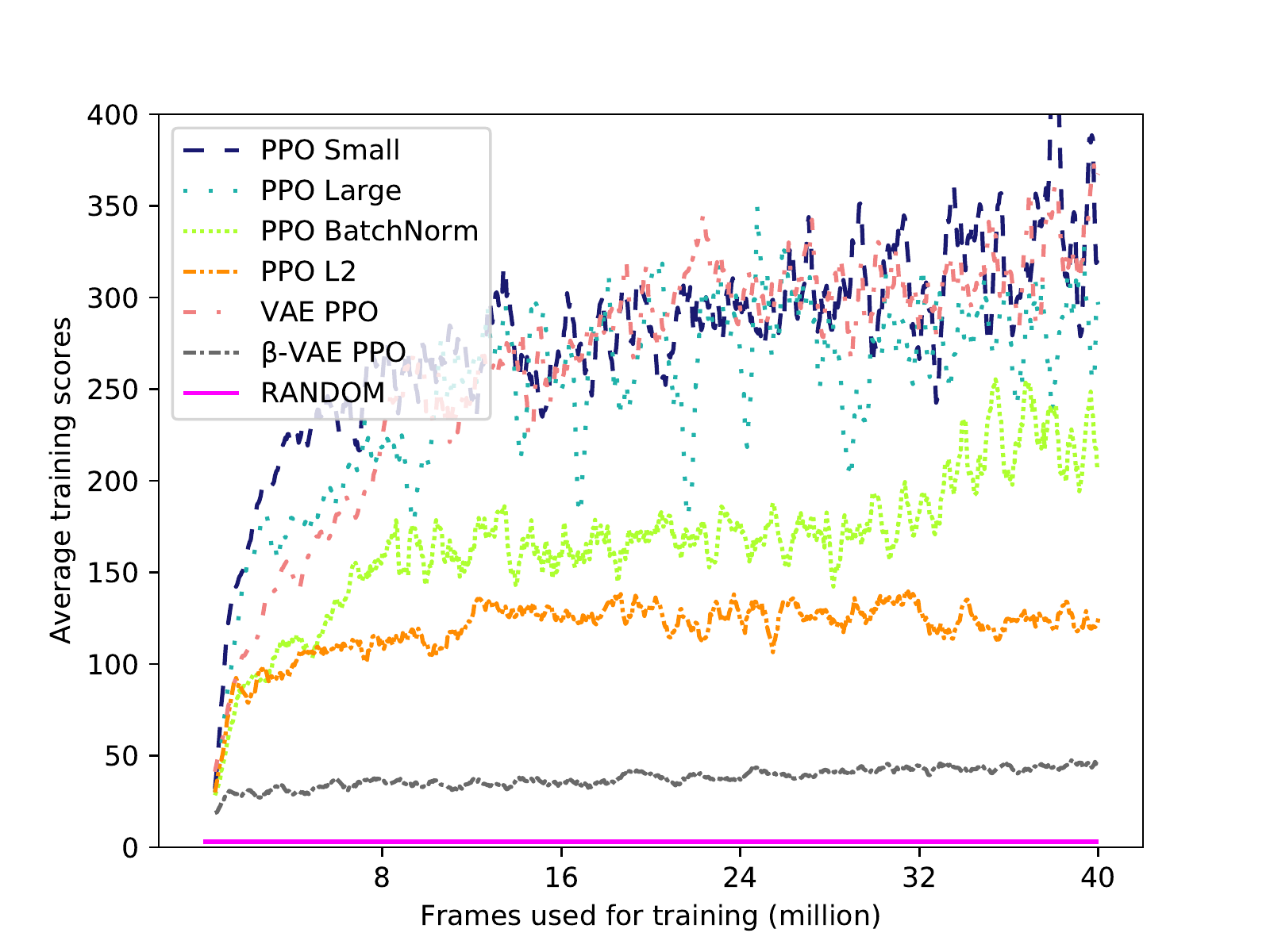}
   \subcaption{Training scores} \label{seed10-t}

   \includegraphics[width=8cm]{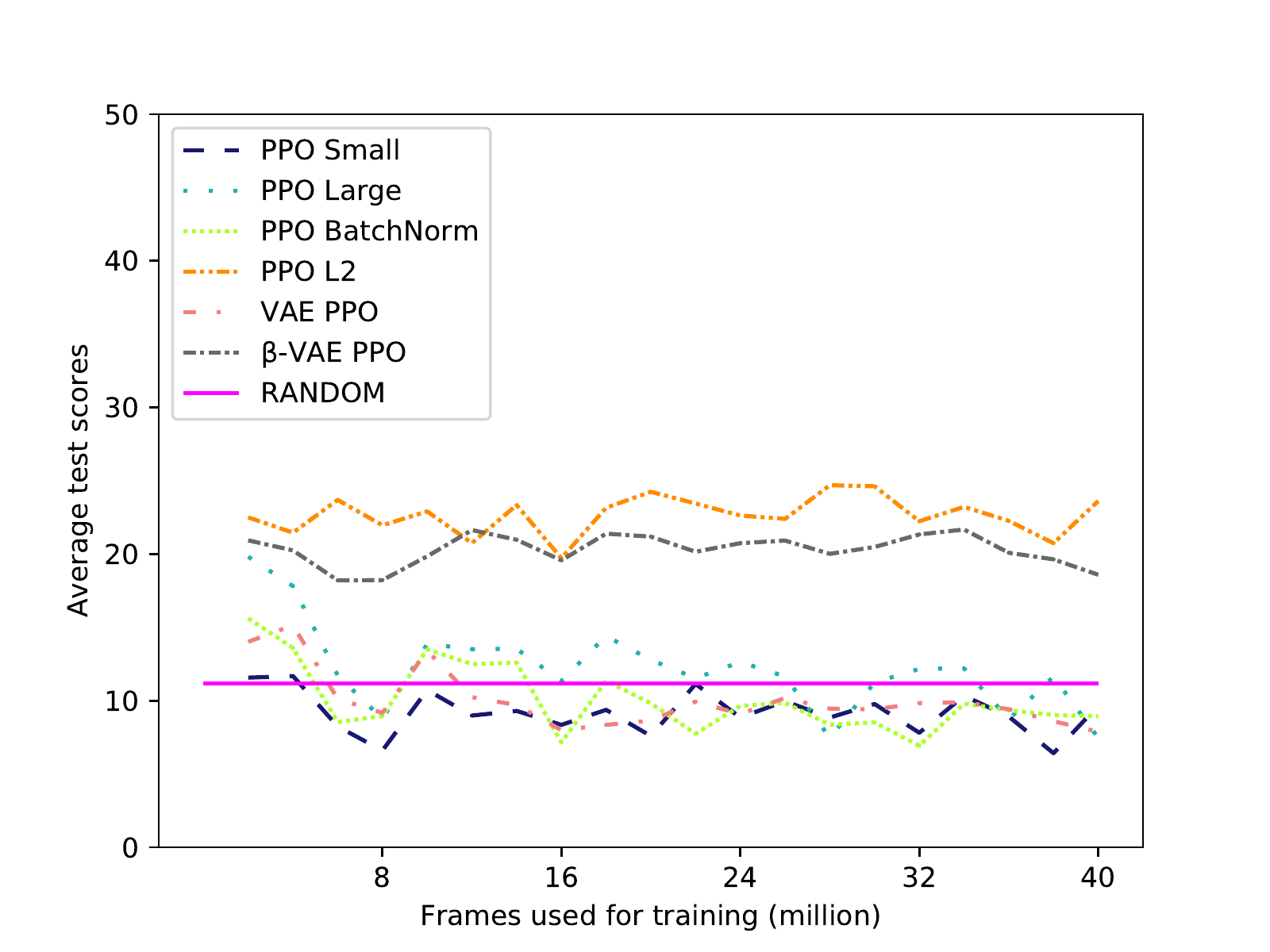}
   \subcaption{Generalization scores} \label{seed10-e}
  \end{minipage}
 \caption{Results with 10 training environments: top figure shows
   training scores and bottom figure shows generalization scores.} \label{seed10}
\end{figure}

We compared the performances of the methods described in the previous
subsection.
Fig.~\ref{seed10} shows the training and generalization scores of these
methods with respect to the number of game frames experienced during
training time.
We used $[0, 9]$ as the set of random seeds of the source environment,
and $[1000,2000)$ of the target.
With respect to the training scores shown in Fig.~\ref{seed10-t}, all
learning methods improved the score much better than the score of random
agents.
We can see that \ppos{} or \textsc{Large}, and VAE PPO performed the
best in the source environment, and \bppo{} performed significantly
worse.

In contrast, Fig.~\ref{seed10-e} shows that generalization scores were
much lower than the training scores for all methods and hardly improved
during the training time.
The most notable result is that of \ppol{}.
It performed slightly better than \ppos{}, which supports the claim by
\cobbe{}. However, the effectiveness of increasing the size of neural
networks was relatively lower than the results reported in the work by
\cobbens{}, in which a larger architecture performed about 20\% better
than a smaller one.
From this result, we can claim that it is less effective in Rogue-Gym to
improve the memorization ability of algorithms, and, thus this benchmark
is fair.
Also, it is surprising that \ppob{} performed worse than \ppos{} and
\ppol{} because it performed better in \cobbens{} in both the training
and test phases.

Only PPO L2 and \bppo{} were clearly better than a random agent, although
their generalization scores were significantly worse than for the
training.
PPO L2 performed the best in the target environment among all methods we
used, which was consistent with the results obtained by \fare{} and
\cobbe{}.
\bppo{} was second-best among all of these methods in the target
environment, while VAE PPO overfitted as did the other methods.
This supports the idea that that disentangled representation learning is
effective for generalization~\cite{DARLA}.

It is notable that all methods other than PPO L2 and  \bppo{}
significantly overfitted.
All methods performed better than the random agent at first but
gradually overfitted and performed worse than the random agent at the
end of the training.
This observation is notable since such large overfitting was not
observed in previous studies like those by \cobbe{} and by \fare{}.

\subsection{Effectiveness of Diversity in Training Environment}
We investigated how an increase in the number of training seeds
(i.e., dungeons) would improve the generalization scores.
In existing works by \zhang{} and \cobbe{}, it is reported that
increasing the number of training levels significantly improves the
performance of generalization.
Accordingly, we conducted experiments with the number of training seeds
set to 20 ($[0, 19]$) or 40 ($[0, 39]$), which is an increase from 10 in
the previous experiments.

In this experiment, we used \bppo{} and PPO L2 since these two methods
performed better than the others in the previous experiment.
In addition, we adopted \ppol{} as the baseline since it has almost the
same number of neural network parameters as these two methods.

\begin{figure}[t]
 \centering
 \includegraphics[width=7.6cm]{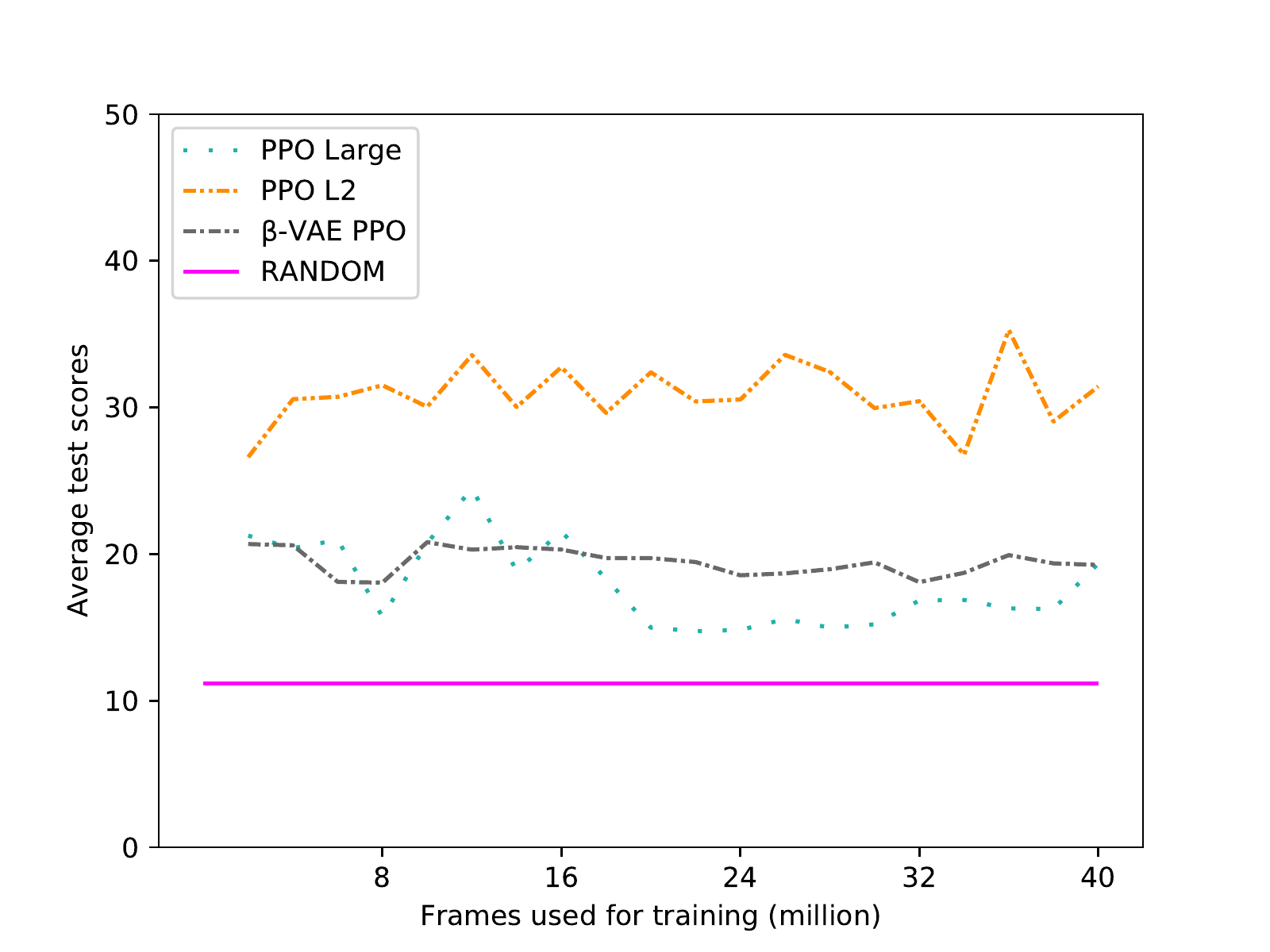}
 \caption{Generalization scores with 20 training seeds} \label{seed20-e}
\end{figure}

\begin{figure}[t]
 \centering
 \includegraphics[width=7.6cm]{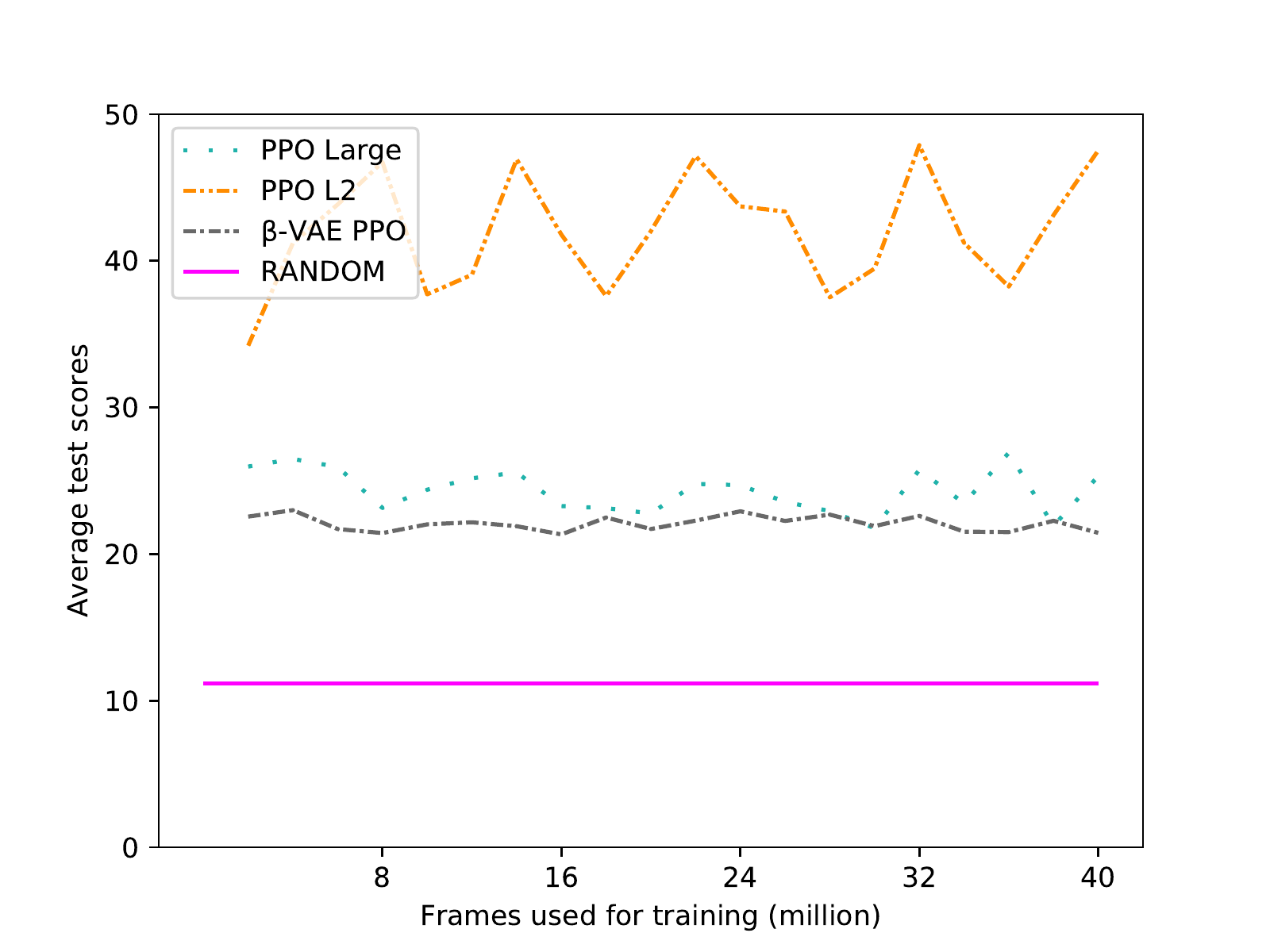}
 \caption{Generalization scores with 40 training seeds} \label{seed40-e}
\end{figure}

Fig.~\ref{seed20-e} and Fig.~\ref{seed40-e} show the results with 20
and 40 training seeds, respectively.
We can see that increasing the number of training seeds improved the
generalization performance for all methods.
However, this improvement was not significant for \bppo{}.
In addition, it is notable that the generalization score of \ppol{}
seemed constant through the training time with 40 training seeds.
This is an improvement since the gradual overfitting
(the decrease in generalization score) observed in the case with 10 or
20 training seeds did not appear here.

\subsection{Effectiveness of Exploration}
Having a policy with higher entropy means that an agent tends to act more
randomly and to explore better.
To investigate the effectiveness of exploration in Rogue-Gym, we ran
agents 10 times for each training seed and measured the entropy of their
policies averaged over ($10\times\text{\#seeds}$) episodes.
\begin{figure}[t]
 \centering
 \includegraphics[width=7.2cm]{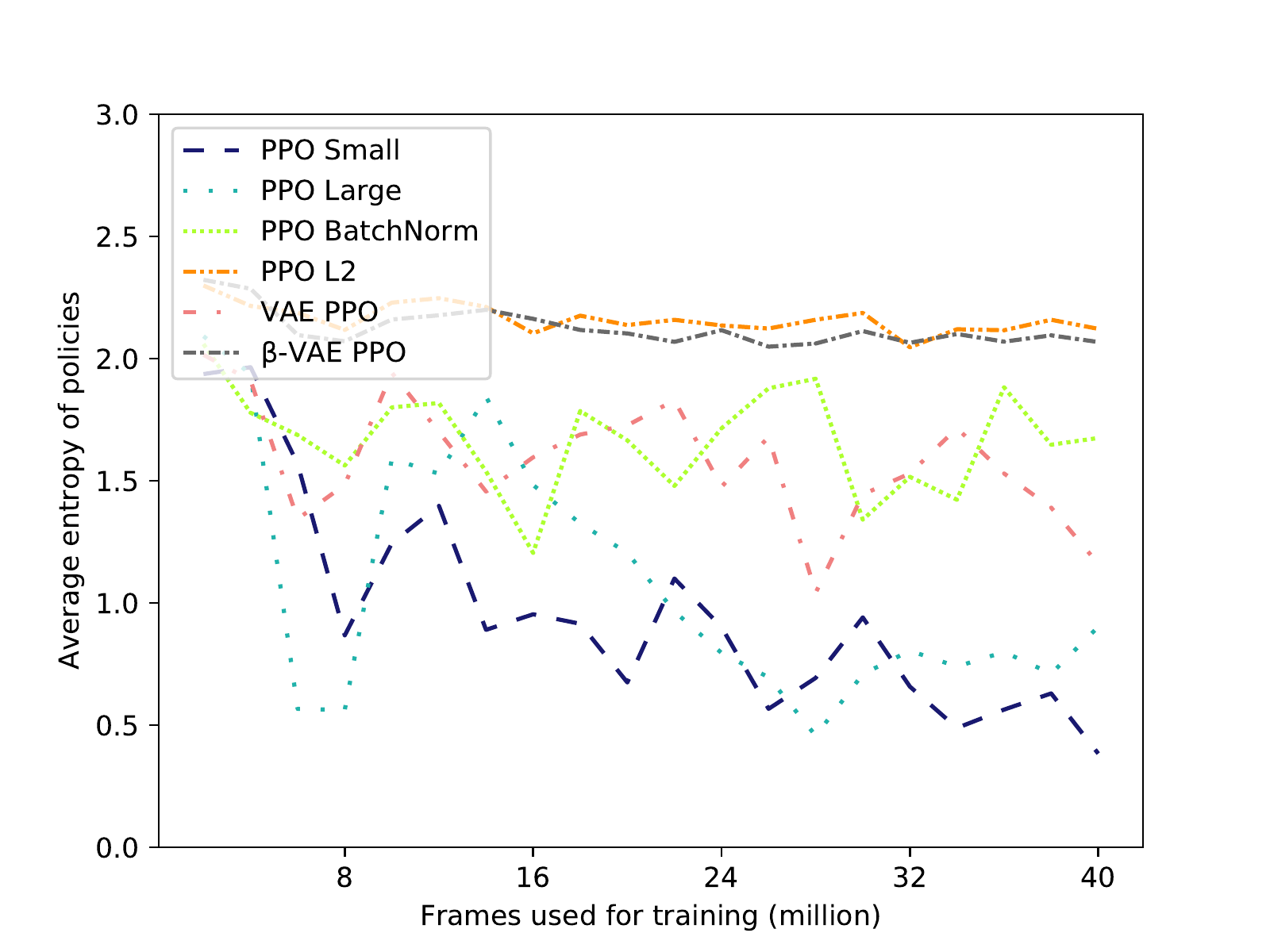}
 \caption{Entropy of policies with 10 training seeds} \label{seed10-en}
\end{figure}

Fig.~\ref{seed10-en} shows the average entropy of policies with 10 training seeds.
We can see that \bppo{} and PPO L2 had policies with higher entropy, and
the entropy of the other methods decreased gradually, corresponding to
the evaluation scores in Fig.~\ref{seed10-e}.
\begin{figure}[t]
 \centering
 \includegraphics[width=7.2cm]{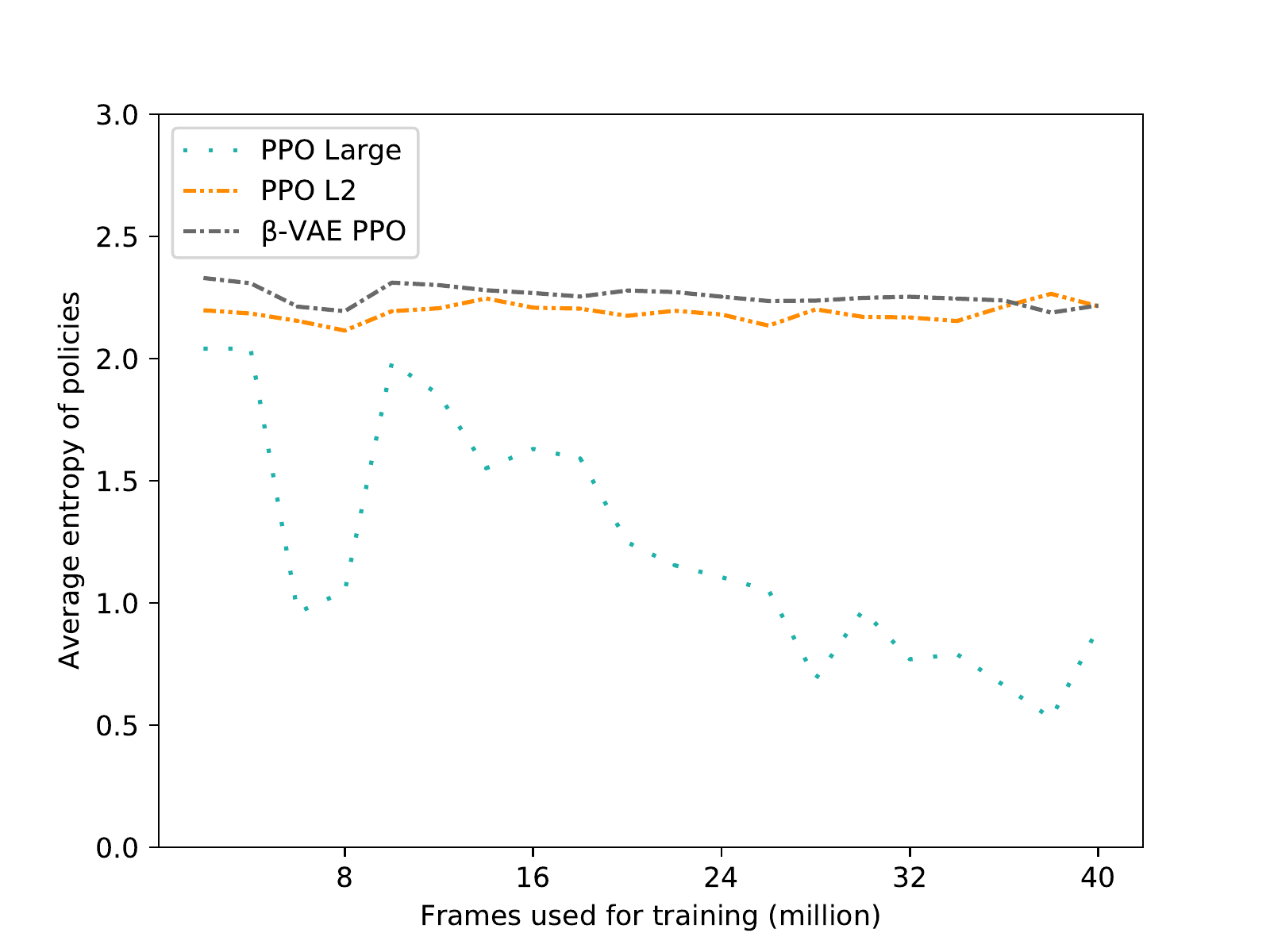}
 \caption{Entropy of policies with 20 training seeds} \label{seed20-en}
\end{figure}
\begin{figure}[t]
 \centering
 \includegraphics[width=7.2cm]{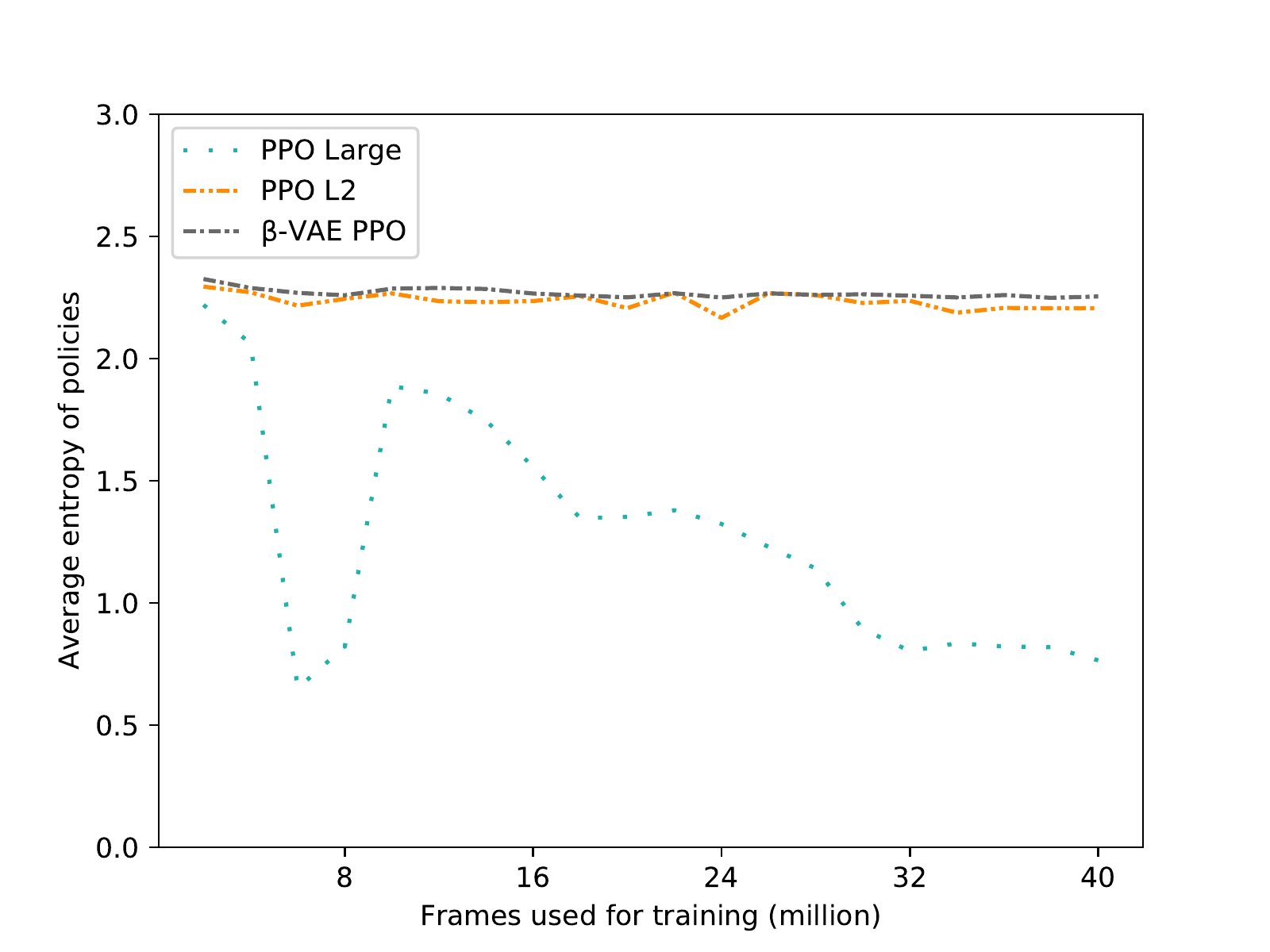}
 \caption{Entropy of policies with 40 training seeds} \label{seed40-en}
\end{figure}

Fig.~\ref{seed20-en} and Fig.~\ref{seed40-en}
show the entropy of policies with 20 and 40 training seeds,
respectively.
We can see that PPO-L2, which had the best generalization score, also had
the highest or similar entropy during training.
Therefore, we can suppose that having diverse policies is important for
obtaining a decent generalization score.
However, the entropy of PPO-L2 policies with 20 or 40 training seeds was
similar, although their generalization scores were different.
Also, the entropy of \bppo{} was similar to that of PPO-L2,  while the
generalization scores significantly differed between the two methods.  
Hence, we can conclude that the randomness of the policies was important
for generalization, but there is certainly another factor that enables
agents to generalize their policies.

\subsection{Detailed Analysis of Learned Policy}
We compared the behaviors of two agents in a specific dungeon to
investigate what kind of generalized behaviors they obtained.
Concretely, we used a \ppol{} agent with 10 training levels as the
representative in overfitted agents and a PPO L2 agent with 40 training
levels as a generalized agent.
To highlight the difference of their behavior, we used the dungeon in
random seed 1008, which is a relatively easier stage.
The scores of these agents were averaged over 10 trials and are summarized
in Table~\ref{s1008}.
\begin{table}[t]
  \centering
 \caption{Generalization scores for seed 1008} \label{s1008}
 \begin{tabular}{l|rD{.}{.}{-1}} \hline
   Method & Number of training levels & \multicolumn{1}{l}{Score} \\ \hline
   \ppol{} & 10 & 0.6 \\
   PPO L2 & 40 & 503.9 \\ \hline
 \end{tabular}
\end{table}

\begin{figure}[t]
 \centering
  \begin{minipage}[cbt]{3.6cm}
   \centering
   \includegraphics[width=3.6cm]{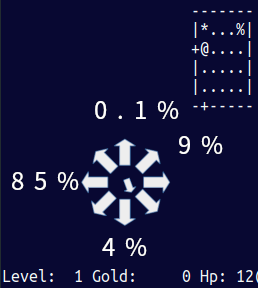}
   \subcaption{Policy for initial state} \label{ppo10-ini}
  \end{minipage}
  \begin{minipage}[cbt]{3.8cm}
   \centering
   \includegraphics[width=3.8cm]{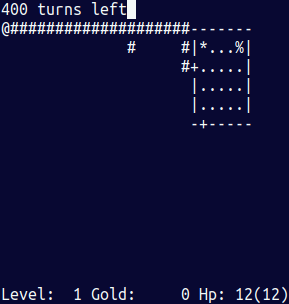}
   \subcaption{After 100 steps} \label{ppo10-100t}
  \end{minipage}
 \caption{Probability of actions of overfitted agent in
   initial state (left) and state after 100 steps (right)} \label{ppo10-act}
\end{figure}

We show the policy for the initial state of the overfitted agent in
Fig.~\ref{ppo10-ini}. 
Each arrow means an action of moving or going down stairs.
The probabilities are shown for the actions of the three highest
probabilities as well as that of moving up.
We expect two actions for generalized agents;
moving up to get gold, or moving right to approach the downstairs.
However, in the policy of the overfitted agent, moving left was the best
action with a high probability of 85\%, and the probability of getting
golds was only 0.1\%.
Moving left was not appropriate because doing so made it more difficult
for the agent to go down to the next floor.
In fact, the agent was not able to go the next level and got stuck in
a passage after some turns passed, as shown in
Fig.~\ref{ppo10-100t}. \par
\begin{figure}[t]
 \centering
  \begin{minipage}[cbt]{3.6cm}
   \centering
   \includegraphics[width=3.6cm]{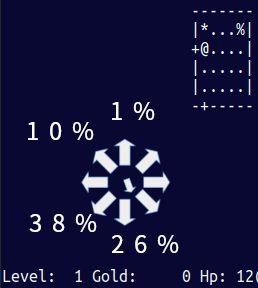}
   \subcaption{Policy for initial state} \label{l240-ini}
  \end{minipage}
  \begin{minipage}[cbt]{3.8cm}
   \centering
   \includegraphics[width=3.8cm]{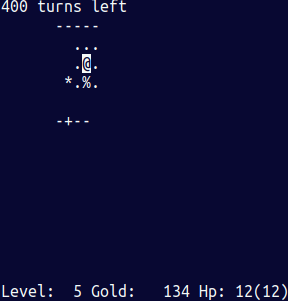}
   \subcaption{After 100 steps} \label{l240-100t}
  \end{minipage}
 \caption{Probability of actions of generalized agent in
   initial state (left) and state after 100 steps (right)} \label{l240-act}
\end{figure}

We show the policy for the initial state of the generalized agent in
Fig.~\ref{l240-ini}.
The probability of getting gold was only 1\%.
It was much higher than that of the overfitted agent but still low.
Also, we can see that this agent tended to move down.
We can not say it was an effective action, but it was better than moving
to the left.
Though we did not see this agent move straight to the right, the
agent often succeeded at going downstairs and reaching deeper levels (5), as
shown in Fig.~\ref{l240-100t}.

\section{Conclusion}\label{section:conclusion}
In this paper, we proposed the Rogue-Gym environment, which is suitable
for researching generalization in reinforcement learning.
Based on roguelike games, it is built with diverse types of dungeons,
fully configurable enemies and items, and has a programming interface
compatible with OpenAI Gym.
Rogue-Gym provides a sufficient number of dungeons by \pcg{}.
Since the dungeons in Rogue-Gym share abstract rules but do not share a
specific structure, it is difficult to get a higher score by memorizing
only good sequences of actions in Rogue-Gym.
Therefore, we believe that the generalization benchmark in Rogue-Gym is fair.

We evaluated existing methods for generalization with PPO in Rogue-Gym.
The difficulty was modestly suppressed by using a small screen size and
by disabling enemies and items so that good policies in the game would
be surely learnable with state-of-the-art methods.
Our results showed that training scores successfully improved during
training, but that generalization scores measured in unseen dungeons
hardly improved and were much lower than the training scores.
Among the generalization methods,  L2 regularization was the most
effective, though there is much room for improvement.
Also, we observed that their entropy of their policies of the agents was
higher for agents with better generalization scores than others.
This suggests that agents need to keep their entropy sufficiently high
to improve their generalization skills. 
The use of maximum entropy reinforcement learning methods like soft
actor-critic~\cite{SAC} would be interesting future work.
Since memory is sometimes useful for handling partial observability
when playing Rogue-Gym, the use of recurrent models (e.g., LSTM) should
also be investigated.

\bibliographystyle{IEEEtran}
\bibliography{IEEEabrv,ref}
\end{document}